\documentclass[runningheads]{llncs}
\usepackage[T1]{fontenc}
\usepackage{graphicx}
\usepackage{amsfonts}
\usepackage{amsmath}
\usepackage{amssymb}
\usepackage{makecell}
\usepackage{tabularx}
\usepackage{multirow}
\usepackage{subcaption}
\newcolumntype{Y}{>{\centering\arraybackslash}X}
\usepackage{orcidlink}

\begin{document}
\title{Empirical investigation of 3D CT Foundation Models and Unsupervised Adaptation for Head and Neck Cancer Recurrence Prediction}
\titlerunning{Empirical investigation of 3D CT Foundation Models}

% \author{Anonymous Authors}
% \authorrunning{A. Author et al.}
% \institute{Anonymous Institution}

\author{Bilel Guetarni\thanks{Corresponding author (\texttt{bilel.guetarni@univ-lille.fr}).}\inst{1}\orcidlink{0009-0003-4340-6158} \and
Feryal Windal\inst{3,4,5,6,7} \and
David Pasquier\inst{1,2} \and
Halim Benhabiles\inst{8}}
\authorrunning{B. Guetarni et al.}
\institute{University of Lille, CRISTAL UMR CNRS 9189, France \and
Academic Department of Radiation Oncology, Centre Oscar Lambret, Lille, France \and
Junia, Lille, France \and
UMR 8520, CNRS, France \and
Centrale Lille, France \and
Univerity of Polytechnique Hauts-de-France, Lille, France \and
University of Lille, France \and
University of Lille, Centre for Digital Systems, IMT Nord Europe, Institut Mines-Télécom, Lille, France}

\maketitle              % typeset the header of the contribution

\begin{abstract}
The rapid emergence of 3D CT foundation models has opened new avenues for predictive modeling from CT imaging, offering a compelling alternative to traditional radiomics which is known to suffer from reproducibility issues and sensitivity to acquisition protocol variations. 
Yet, as these models grow in availability, a critical need arises to evaluate how well their learned representations generalize across diverse clinical settings and whether adaptation to specific downstream tasks is necessary to unlock their full potential. 
To address these questions, we benchmarked several 3D CT foundation models for predicting recurrence-free survival in head and neck cancer across two public datasets totaling 3,644 patients, evaluating various adaptation strategies and modality fusion mechanisms. 
Our findings reveal persistent difficulty in identifying features that generalize consistently across different imaging distributions, as evidenced by significant performance drops on external validation cohorts. 
Ultimately, the integration of imaging features with clinical data remains the most accurate approach for prognostic prediction, though achieving universal generalization across varied clinical contexts continues to represent a substantial challenge for the current generation of models.
\end{abstract}

\section{Introduction}
%    - transfer strategy (freeze or fine-tune ?)
%    - localisation specific demographics? (tests on separate groups of RADCURE localisation)

Head and neck cancer is a major public health concern, accounting for more than 1.7 million new cases and over 500,000 deaths annually \cite{sun2025global}.
Despite advances in multimodal therapies, disease recurrence remains frequent, with over 50\% of patients experiencing relapse within two years after initial treatment \cite{sun2025global}. 
This high rate of recurrence substantially compromises overall survival and underscores the urgent need for more effective risk stratification, as well as adaptive treatment and follow-up strategies.

In this context, developing reliable recurrence prognostic tools is essential to support clinicians in identifying high-risk patients and enabling more personalized treatment.
In this sense, patients may benefit from treatment escalation \cite{ruhle2020value} or de-escalation which may reduce risks of radiotherapy-induced toxicities \cite{adelstein2019role}.
As 3D Computed Tomography imaging (CT scans) is routinely used to visualize head and neck tumors and for post-treatment follow-up, it may provide a relevant modality for recurrence prediction. 
Moreover, its integration with patient clinical information could further enhance predictive performance.

A prominent line of research for characterizing these CT scans involves radiomics, which utilizes mathematically defined descriptors to characterize the shape, intensity, and texture of regions-of-interest.
While radiomics provides a structured framework for analyzing medical images, it is hindered by significant reproducibility issues; indeed radiomics are known to be protocol-specific and not generalize across diverse imaging protocols and reconstruction kernels \cite{zhao2016reproducibility,lambin2017radiomics}.
In particular, a recent study showed a difficulty in reproducing associations between radiomics and clincial endpoints identified in a prior work \cite{berger2023assessing}.
In parallel, an alternative approach has emerged through the development of 3D vision foundation models (VFMs) that are pre-trained on large quantities of CT scans.
Unlike traditional methods, these models leverage unsupervised pretext tasks to learn general invariant features directly from the data distribution \cite{pai2025vision}.
By capturing the latent structure of volumetric data without the need of annotations, these models generate high-level representations offering a general characterization of complex morphological patterns. \\

Despite the availability of these 3D CT foundation models, their utility in prognostic tasks depends heavily on how they are integrated into downstream pipelines. 
When these models are employed as frozen feature extractors, the primary challenge shifts from the pre-training itself to the design and adaptation of the subsequent predictor. 
There is currently a lack of systematic studies addressing how to map these high-dimensional latent embeddings to specific clinical outcomes, particularly regarding how to fuse them with complementary modalities or optimize the classification head.
Consequently, there is a need for comprehensive benchmarks permitting to compare different 3D CT foundation models and evaluate various approaches for adapting downstream predictors to these frozen backbones.
Such a framework is essential for moving toward a more structured understanding of how to leverage large-scale pre-trained models for reliable prognostic tasks in oncology. \\

In this work, we address the aforementioned questions by benchmarking several state-of-the-art VFMs on the task of recurrence prediction at two years in patients treated with radiotherapy in a curative intent for head and neck cancer.
We intentionally exclude traditional radiomics from this benchmark, as our primary objective is an empirical investigation of CT foundation models.
We have systematically investigated different strategies for training the predictor, as well as multiple approaches for multimodal data fusion combining CT imaging and clinical variables (including dosimetric information).
The performance evaluation of the predictors has been conducted using a comprehensive set of metrics to ensure robust and reliable comparisons. 
All experiments have been performed on a large-scale public dataset, namely RADCURE \cite{welch2023computed,welch2024radcure}, 
comprising 3,346 patients, and all classifiers 
have been further validated on an external public dataset, Head-Neck-PET-CT \cite{vallieres2017radiomics}, comprising 298 patients, to evaluate their 
potential for generalizability.
For reproducibility purposes, the code and extracted VFM features will be made publicly available upon publication.

\section{Related Works}
% Works that validate on external cohorts [Bae et al. 2024, Le at al. 2022]
% inclure Diamant et al. Mateus et al. Vallieres et al.

Most studies on recurrence predictive models employ radiomics as imaging features due to their genericity and easy computation simplified by standardized libraries (e.g., PyRadiomics) \cite{haider2021prediction,teng2022improving,lv2022context,shannon2025leveraging}.
Applied to CT volumes with delineated regions-of-interest, they can be used to build risk stratification models using for example Cox's proportional hazards \cite{lv2022context} or random forests \cite{haider2021prediction}.
Specific time endpoints models are also widely used (e.g., recurrence at 2 or 5 years) \cite{shannon2025leveraging,teng2022improving} as they have the advantage of fixing a particular point in time for the event in question. \\

More recently, for the reproducibility reasons stated earlier \cite{zhao2016reproducibility,lambin2017radiomics}, several studies have leveraged deep learning-based imaging characterization driven by the availability of public datasets such as RADCURE \cite{welch2024radcure}.
These studies employ diverse types of end-to-end deep learning architectures such as Graph Neural Networks \cite{bae2024hog} or Convolutional Neural Networks \cite{le2022cross}.
Although these methods achieve strong performances, they often require computationally heavy hyperparameters tuning steps and can fail to generalize to external cohorts with different imaging features distribution due to misalignment in acquisition protocols.
VFMs have been proposed as a solution to these challenges as they do not require any hyperparameter tuning once trained, and are pre-trained on large heterogeneous datasets covering many centers in a task agnostic manner \cite{pai2025vision,hamamci2026generalist,he2025vista3d,li2025well}.
Pai et al. \cite{pai2025vision} pre-trained a SegResNet on a self-supervised contrastive task to generate similar latent embeddings for two augmented versions of the same patch.
They propose a variant of the SimCLR task \cite{chen2020simple} to guarantee that positive and negative pairs are different.
The model showed strong results on several downstream task scenarios as organ segmentation, tumor segmentation and head abnormality classification.
The authors also found the model to be robust under perturbations.
Similarly, Hamamci et al. \cite{hamamci2026generalist} performed contrastive self-supervised learning to train a combination of language-image model composed of a text and an image encoders.
These are jointly optimized to generate similar latent embedding for pairs of text and image extracted from a curated dataset collected beforehand.
When evaluated, the image encoder outperforms fully supervised approaches on multi-abnormality detection and case retrieval.
Li et al. \cite{li2025well} pre-trained diverse deep learning architectures for segmentation of 25 anatomical structures on a detailed voxel-level annotated dataset.
They show that their trained models have stronger transfer abilities to new 3D segmentation tasks than existing models, especially on classes with limited training samples.
He et al. \cite{he2025vista3d} proposed a foundation model for 3D CT segmentation that supports 127 different structures including organs, bones and arteries.
Their interactive training framework allows to generate, during the training, new annotations on unlabelled images by allowing human annotators to be solicited.
Their model was shown to be competitive with specialized models on the full range of supported classes, as well as, outperforming state-of-the-art methods in zero-shot transfer tasks. \\

Fine-tuning these large-scale models comes with a substantial computational cost.
Utilizing them as feature extractors provides a more practical alternative, similar to radiomics.
In this setup, only the subsequent, smaller predictor is trained, often from scratch, to map the VFM features to the target outcome.
A significant challenge remains, however, as the limited availability of labeled data may be insufficient for this downstream classifier to effectively adapt to the feature space of the foundation model.
Indeed, such constraints make it challenging to adapt to never seen before categories especially in the context of deep neural networks, prone to overfitting.
To fix this issue, few-shot learning is a paradigm in which a classifier is accommodated to new classes with only a few samples at hand.
In this context, Snell et al. \cite{snell2017prototypical} proposed prototypical networks (ProtoNet), in which a model is optimized to generate a metric space inside which new samples can be classified by computing the distance towards prototypes (i.e., class centroids).
To this end the model is trained, from a few number of samples, to minimize the distance between the samples of the same class and their centroid, and maximize the distance with respect to all other classes centroids.
Comparative studies on natural images datasets demonstrated the advantage of ProtoNet over other competitive methods of the literature.
Hu et al. \cite{hu2022pushing} proposed an extension of this approach by incorporating it into a three stages process including a pre-training stage with self-supervised contrastive tasks.
Their experiments suggest a positive margin of downstream performance compared to previous methods. \\

The promising results demonstrated by these diverse foundation models suggest they could provide a more robust alternative to traditional radiomics for clinical applications. 
However, their specific utility for predicting oncological outcomes and the most effective way to adapt their features to such tasks remains to be fully established. 
This highlight the need to systematically benchmark these models to evaluate their predictive potential and generalization capabilities, including in the context of head and neck cancer recurrence.

\section{Foundation model-based prognosis prediction}

\begin{figure}
   \includegraphics[width=\textwidth]{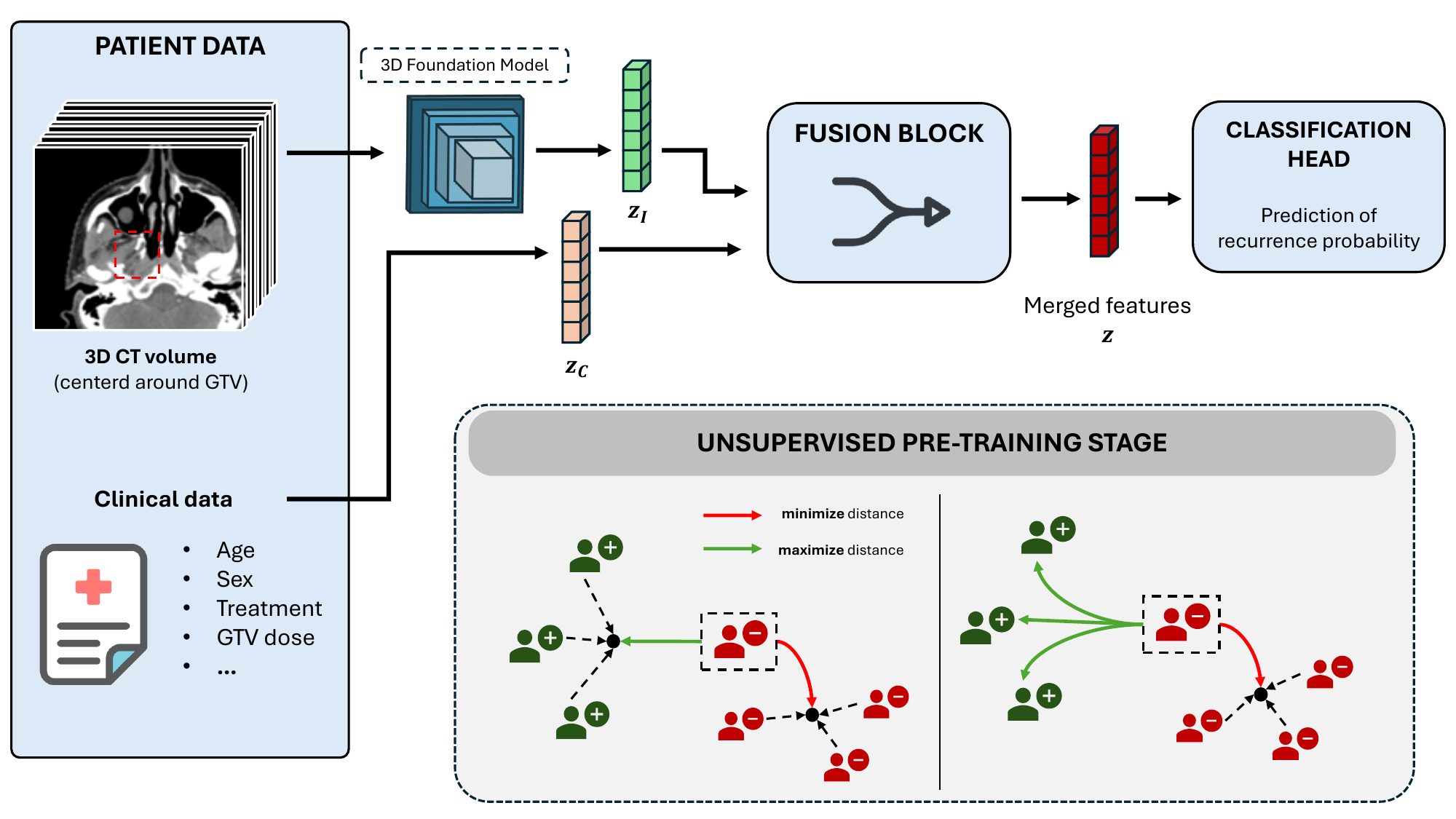}
   \caption{Proposed empirical investigation of combining Vision Foundation Model and unsupervised adaptation for recurrence prediction. 
   In the bottom grey box we illustrate the difference between ProtoNet (\textbf{left}) and our proposed variant (\textbf{right}).}
   \label{pipeline}
\end{figure}

As illustrated in Figure \ref{pipeline}, we study three key aspects of VFM-based approaches that build prognostic predictive models.
We study the impact of the foundation models conception on the features relevance for the prediction task.
This includes the model architecture, pre-training objective, as well as the data quantity and diversity.
The second element concerns how the imaging features and clinical data are fused together.
As various strategies exist and selecting the most appropriate one is not trivial, we compare different approaches.
The third element is the most efficient way for training the predictor from the modalities features.

\subsection{Impact of foundation model conception on features transfer}
We selected four 3D CT VFMs from the literature, chosen to cover different pre-training strategies, model size, as well as different training data quantity and source.
CT-FM \cite{pai2025vision} is a 77M parameters SegResNet encoder pre-trained on 148K CT scans with contrastive self-supervised learning.
CT-CLIP \cite{hamamci2026generalist} is an image-language model pre-trained using contrastive image-language pre-training on 25K pairs of scans and textual reports.
We only use the 25M parameters image encoder consisting of a 3D vision transformer.
VISTA3D \cite{he2025vista3d} is a SegResNet encoder trained on 127 anatomical classes for 3D CT segmentation.
The authors curated a dataset of 11K 3D CT scans to train the model.
SUPREM \cite{li2025well} is a suite of pre-trained 3D segmentation models covering 25 anatomical structures trained on 9K CT scans; we employ the 4.7M parameters SegResNet version published by the authors in our experiments. \\

In this work, VFMs are used as feature extractors of the CT volume, centered around the gross tumor volume (GTV).
More precisely, expert annotation of the GTV is used to center crop the CT, while respecting the input size requirements of each VFM.
For segmentation models, the output of the encoder's last layer is converted into an embedding vector through a global average pooling.
For the other models, the same operation is applied on their last layer.

\subsection{Modality fusion functions}
\label{section-modality-fusion}

Once imaging features are extrated by the VFM, they are combined with clinical data to produce a multimodal patient representation.
The choice of this fusion function directly affects how much each modality contributes to the final prediction.
We compare three fusion mechanisms.

\subsubsection{Concatenation} Imaging and clinical features are projected to a common embedding dimension by a linear projection.
The projected features are then concatenated followed by a 2-layer MLP with GELU activation and dropout.

\subsubsection{Gating} Inspired by the gating mechanism used in recurrent neural networks \cite{cho2014learning}, we introduce weighting factors that modulate the contribution of each modality.
Similar to concatenation, imaging and clinical features are first projected to a common embedding followed by concatenation and batch normalization.
The concatenated vector is then used to compute gating scores through a linear layer, which are then used to weight the modalities features vectors in the fused vector.
Finally we apply a 2-layer MLP with GELU activation and dropout.

\subsubsection{Attention} Finally, we consider the attention mechanism from the Transformer architecture \cite{vaswani2017attention}.
The modalities features vectors are projected to a common latent embedding dimension, and we compute attention scores using a learnable query vector (initialized randomly before training).
The fused vector is the weighted sum of the modalities vectors using these scores.
We also apply a 2-layer MLP with GELU activation and dropout to the fused vector.

\subsection{Predictor training strategies}

The overall predictor, consisting of the fusion module and the classification head, is trained in two different ways.
First, we train the predictor from scratch in a classical supervised setting.
Second, we propose an unsupervised adaptation of the predictor to the VFM features.
This is implemented as a pre-training objective that optimizes the fusion module to cluster patients according to their prognostic class.
This is due to the fact that the available patient cohorts may be insufficient for a downstream predictor to naturally adapt to frozen foundation model embeddings.
We therefore employ an unsupervised adaptation step to align the fusion module with the latent imaging representations. \\

More specifically, for the second training strategy we base our work on the few-shot learning paradigm \cite{snell2017prototypical}.
In a few-shot learning setting, a classifier is trained to learn new classes with a few samples of each class.
This can be challenging and lead to overfitting, especially with high-dimensional features.
For example, Snell et al. \cite{snell2017prototypical} proposed ProtoNet to train a model by optimizing its latent embedding such that samples belonging to the same class are close to one another (i.e., close to the class centroid).
In ProtoNet, one is given a small support of samples belonging to two classes\footnote{this is specific to our downstream task, the approach is generic to any number of classes and we refer the reader to the original paper.}, $S=\{(x_i, y_i) \vert i=1...N \}$, with $x \in \mathbb{R}^{D}$ and $y \in \{0,1\}$.
$S^+$ is the subset of positives samples ($y=1$), and similarly, $S^-$ the set of negative ones ($y=0$).
For a sample belonging to the positive class, the training objective consists of minimizing the distance with the class centroid defined as:
\begin{equation}
   c^+ = \frac{1}{\vert S^+ \vert}\sum_{(x_i,y_i) \in S^+}f_\theta(x_i)
\end{equation}
The negative class prototype $c^-$ is defined equivalently.
The model $f_\theta : \mathbb{R}^D \to \mathbb{R}^M$ is trained to maximize the likelihood of a sample belonging to its class.
In the case of a positive sample ($y=1$), this can be formulated as:
\begin{equation}
   p_\theta(y=1|x) = \frac{exp(-d(f_\theta(x),c^+))}{exp(-d(f_\theta(x),c^+)) + exp(-d(f_\theta(x),c^-))}
   \label{equation-protonet}
\end{equation}
where $d$ is a distance function. \\

In the standard ProtoNet, a sample is compared to the centroid of the opposing class (see denominator of Eq. \ref{equation-protonet}).
We investigate whether it is better to compare it to each of its individual samples. 
We propose a variant, inspired by the Cox survival model that maximizes the partial likelihood, i.e., the probability of observing an event for a given sample relative to all others.
Following this, we modify the class-level distance computation with an instance-level one for the opposite class:
\begin{equation}
   p_\theta(y=1|x) = \frac{exp(-d(f_\theta(x),c^{+}))}{exp(-d(f_\theta(x),c^{+})) + \sum_{(x_{i}',y_i) \in S^{-}}{exp(-d(f_\theta(x),f_\theta(x_{i}')))}}
\end{equation}
In both cases, the loss function is the negative log-likelihood which is defined as: $-\log{p_\theta(y|x)}$. \\

We qualify this method as unsupervised since the patients labels are not directly used in the loss formulation.
Instead, the class is used to group patients together for computing the prototypes and samples distances.

\section{Experiments}
% p-value between pos and neg groups using TNM stage and tumor volume as variables.
% SHAP values for clinical features (verify if TNM stage most important one).

\subsubsection{Datasets.} To perform our study we employed two publicly available datasets: RADCURE \cite{welch2023computed,welch2024radcure} and Head-Neck-PET-CT (HN-PETCT) \cite{vallieres2017radiomics}.
Each patient has a planning CT, from which we extract VFM features, and its associated clinical data which includes: age, sex, treatment type (radiotherapy alone or combined with chemotherapy) and the TNM stage.
We also include the GTV total dose computed as the median voxel value in the dose maps using the GTV contours provided with the CT.
Age and dose are normalized and categorical variables one-hot encoded.
To avoid under-represented clinical categories, we apply the following inclusion criteria: no surgery prior to radiotherapy, non metastatic (\textbf{M0}) oropharyngeal cancer.
We studied 2-year recurrence as the clinical endpoint of prediction.
This endpoint was selected because most recurrences in head and neck cancer occur during the first two years following treatment \cite{sun2025global}
Survival data are binarized accordingly: patients with an event before the endpoint are labeled positive, others negative. 
Patients with last follow-up before the endpoint and no observed event are censored and therefore excluded of the study.
For RADCURE, we use the train/test split provided in the TCIA repository \cite{welch2023computed} after applying the previously mentioned inclusion criteria.
Training patients are used for hyperparameters search and test patients are held out for evaluation.
HN-PETCT is exclusively used as an external validation dataset.
In total, we have 1,554 training, 499 validation and 532 testing samples.

\subsubsection{Evaluation metrics.}
We report three classification metrics: AUC, F1-score and balanced accuracy (BA, the average of specificity and sensitivity).
AUC and F1-score are the most common metrics in healthcare literature \cite{collins2024evaluation}.
BA is included because survival datasets can become highly imbalanced once binarized: 25\% of RADCURE and 20\% of HN-PETCT patients are positive.
Therefore, selecting appropriate metrics that are not impacted by class imbalance is necessary to perform clinically relevant comparisons.
BA is robust to this imbalance and offers a good trade-off between sensitivity and specificity.
Reported metrics are averaged over 10 bootstraps.

\subsubsection{Training setup.}
As stated earlier, the VFMs are used as feature extractors and are therefore, not trained, unlike the fusion module and classification head.
Unless stated otherwise, all models were trained for 200 epochs with a batch size of 16, an Adam optimizer with learning rate of $5 \times 10^{-5}$ and dropout ($p=0.5$).
To counter class imbalance, we apply random undersampling on the majority class to match the size of the minority one.
For unsupervised pre-training, models are trained for 2000 steps with a batch size of 128 using Adam with a cosine learning rate schedule: initialized at $10^{-6}$, warmed-up over 50 steps to $5 \times 10^{-5}$ then decayed back to $10^{-6}$.
L2 weights decay ($\lambda=0.1$) is applied in both stages and we use binary cross-entropy as the loss function.

\section{Results}

\begin{table}[!t]
   % from 106 taking best results for each VFM and pre-training method (all modalities and fusion functions included)
   \centering
   \caption{Comparison of different pre-training strategies and Vision Foundation Models.}
   \label{compare-pretrain-strategies}
   \begin{tabularx}{\textwidth}{|l|l|c|YYY|YYY|}
      \cline{4-9}
      \multicolumn{1}{l}{} & \multicolumn{1}{l}{} & \multicolumn{1}{c}{} & \multicolumn{3}{|c|}{\textbf{RADCURE}} & \multicolumn{3}{c|}{\textbf{HN-PETCT}} \\
      \hline
      VFM & pre-training & multimodal & AUC & BA & F1 & AUC & BA & F1 \\
      \hline
      \multirow{3}{*}{CT-CLIP} & none & \checkmark & 0.625 & 0.591 & 0.390 & 0.641 & \textbf{0.600} & 0.391 \\
      & ProtoNet & \checkmark & 0.589 & 0.502 & 0.287 & 0.573 & 0.503 & 0.275 \\
      & Cox-like ProtoNet & \checkmark & 0.637 & 0.596 & 0.437 & \textbf{0.658} & 0.582 & \textbf{0.406} \\
      \hline
      \multirow{3}{*}{SUPREM} & none & \checkmark & 0.596 & 0.528 & 0.222 & 0.572 & 0.507 & 0.087 \\
      & ProtoNet & \checkmark & 0.619 & 0.508 & 0.308 & 0.584 & 0.500 & 0.229 \\
      & Cox-like ProtoNet & \checkmark & 0.607 & 0.574 & 0.410 & 0.485 & 0.500 & 0.013 \\
      \hline
      \multirow{3}{*}{CT-FM} & none & \checkmark & 0.643 & 0.611 & 0.442 & 0.539 & 0.515 & 0.349 \\
      & ProtoNet & \checkmark & 0.655 & 0.617 & 0.451 & 0.500 & 0.491 & 0.338 \\
      & Cox-like ProtoNet & \checkmark & 0.647 & 0.613 & 0.440 & 0.541 & 0.526 & 0.348 \\
      \hline
      \multirow{3}{*}{VISTA3D} & none &  & \textbf{0.668} & \textbf{0.638} & \textbf{0.476} & 0.545 & 0.518 & 0.305 \\
      & ProtoNet &  & 0.661 & 0.627 & 0.463 & 0.545 & 0.528 & 0.315 \\
      & Cox-like ProtoNet &  & 0.662 & 0.627 & 0.463 & 0.560 & 0.530 & 0.317 \\
      \hline
   \end{tabularx}
\end{table}

\subsection{Foundation model embedding}
\label{section-fm-embed}

Tables \ref{compare-pretrain-strategies} reports the performance of the four VFMs under different pre-training strategies.
On RADCURE, VISTA3D features clearly outperform the other VFMs, particularly when the predictor is trained without pre-training (AUC 0.668, BA 0.638 and F1 0.476). 
Notably, VISTA3D reaches its best performance when imaging features are used alone: adding clinical data decreases performance by 0.029 AUC, 0.026 BA and 0.029 F1. \\

However, when evaluated on the external cohort, we observe a significant drop of performance across every training strategies (i.e., with or without pre-training).
In contrast, CT-CLIP displays better generalization capability than other VFM: combined with our Cox-like ProtoNet pre-training, it reaches an AUC of 0.658, 0.582 BA and 0.406 F1.
This indicates that CT-CLIP features are more robust and can better generalize to other acquisition protocols.
Indeed, the difference of AUC between the internal and external cohorts is 0.021 for CT-CLIP, 0.035 for SUPREM, 0.123 for VISTA3D and 0.155 for CT-FM.
Robustness to data distribution shifts is, indeed, a characteristic expected from foundation model embeddings.
We only found that CT-CLIP exhibits such robustness across all considered metrics. \\

Nonetheless, the reported classification performances are not sufficient to argue in favor of VFMs as ready-to-use embedding models for recurrence prediction, specifically with limited training data.
We observe that most AUCs on the external cohorts are below 0.6 and every model shows an F-score lower than 0.5.

\subsection{Contribution of multimodality for recurrence prediction}

\begin{table}[!t]
   % from 106 taking best results for each VFM and pre-training method (all modalities and fusion functions included)
   \centering
   \caption{Comparison of fusion functions and unsupervised adaptation. We use CT-CLIP for imaging features.}
   \label{compare-fusion-functions}
   \begin{tabularx}{\textwidth}{|l|c|YYY|YYY|}
      \cline{3-8}
      \multicolumn{1}{l}{} & \multicolumn{1}{c}{} & \multicolumn{3}{|c|}{\textbf{RADCURE}} & \multicolumn{3}{c|}{\textbf{HN-PETCT}} \\
      \hline
      modality & fusion function & AUC & BA & F1 & AUC & BA & F1 \\
      \hline\hline
      \multicolumn{8}{|c|}{w/o pre-training} \\
      \hline
      clinical & & 0.553 & 0.527 & 0.293 & 0.552 & 0.530 & 0.291 \\
      image & & 0.519 & 0.502 & 0.132 & 0.484 & 0.509 & 0.143 \\
      multi & concat & 0.594 & 0.501 & 0.219 & 0.593 & 0.508 & 0.244 \\
      multi & gating & 0.625 & 0.591 & 0.390 & 0.641 & \textbf{0.600} & 0.391 \\
      multi & attention & 0.560 & 0.510 & 0.152 & 0.579 & 0.511 & 0.154 \\
      \hline\hline
      \multicolumn{8}{|c|}{Cox-like ProtoNet} \\
      \hline
      clinical & & 0.587 & 0.520 & 0.275 & 0.575 & 0.507 & 0.218 \\
      image & & 0.506 & 0.503 & 0.274 & 0.505 & 0.502 & 0.232 \\
      multi & concat & 0.590 & 0.500 & 0.082 & 0.580 & 0.500 & 0.078 \\
      multi & gating & \textbf{0.637} & \textbf{0.596} & \textbf{0.437} & \textbf{0.658} & 0.582 & \textbf{0.406} \\
      multi & attention & 0.527 & 0.520 & 0.310 & 0.563 & 0.539 & 0.354 \\
      \hline
   \end{tabularx}
\end{table}

\begin{figure}[!b]
   \includegraphics[width=\textwidth]{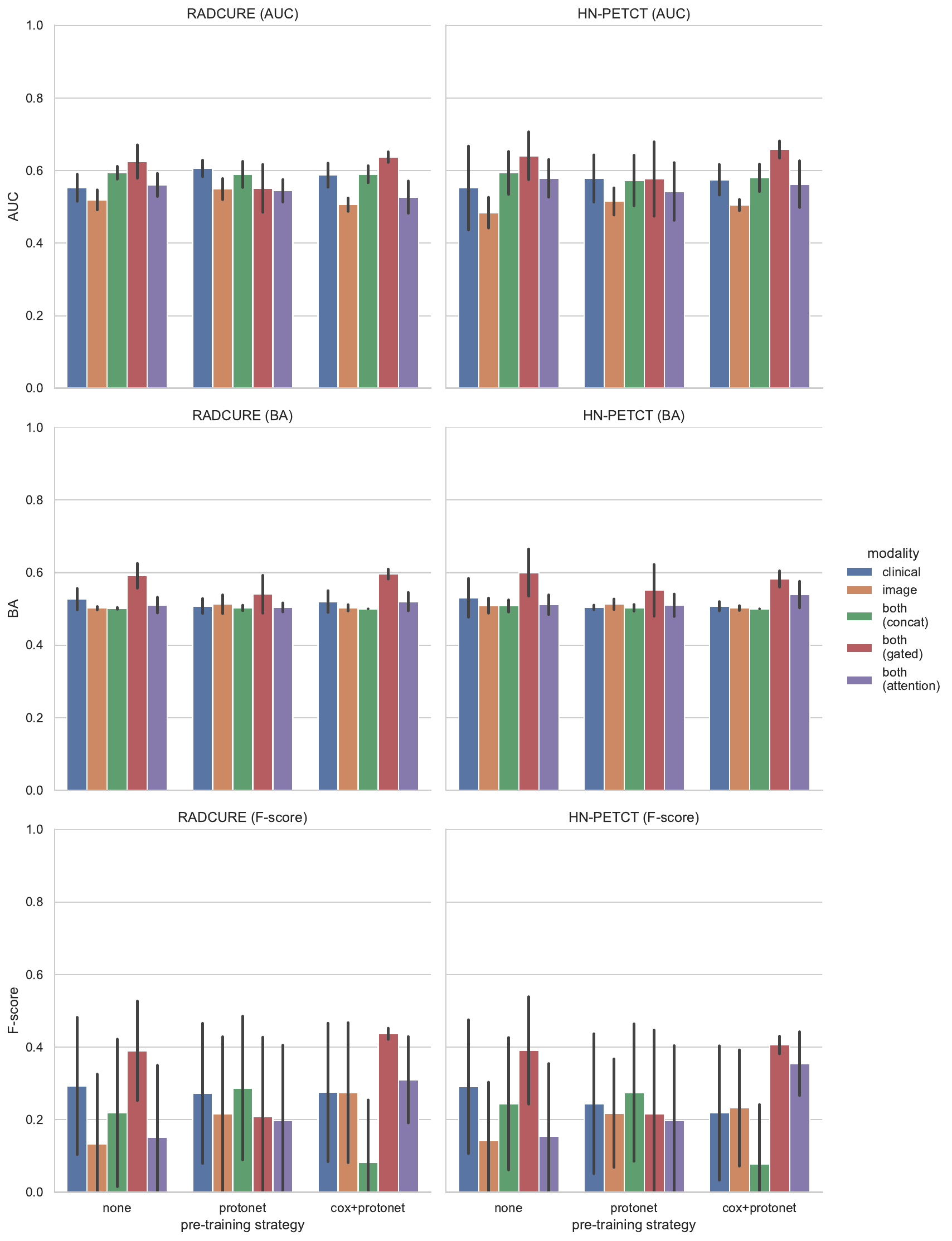}
   \caption{AUC, BA and F-score reported on both RADCURE and HN-PETCT using CT-CLIP for imaging features.}
   \label{modality-pretraining}
\end{figure}

Since CT-CLIP showed the highest generalization capability on the external cohort (see section \ref{section-fm-embed}), we use it with our Cox-like ProtoNet pre-training as the base configuration to compare the three fusion functions described in section \ref{section-modality-fusion}.
Results are reported in Table \ref{compare-fusion-functions} and Figure \ref{modality-pretraining}.
We can observe that gating is clearly the most effective fusion function on both cohorts, with an AUC of 0.637 on RADCURE and 0.658 on HN-PETCT.
It outperforms both concatenation (0.590 and 0.580) and attention (0.527 and 0.563).
Moreover, we can note that imaging features only do not compete with clinical data that consistently perform better across the all metrics and datasets.

\subsection{Pre-training to improve generalization}

Table \ref{compare-pretrain-strategies} shows that unsupervised pre-training improves external generalization on both RADCURE and HN-PETCT for most VFMs.
On RADCURE, pre-training also improves performance for the majority of them, with the exception of VISTA3D whose features are already well-suited to the internal cohort without adaptation. 
Our Cox-like ProtoNet variant is the best pre-training strategy in three out of four cases.
On HN-PETCT, for CT-CLIP, VISTA3D, and SUPREM the best AUC is achieved with pre-training, with respective gains of +0.017, +0.015, and +0.012 compared to no pre-training. 
CT-FM shows benefit of pre-training in terms of AUC (+0.017) and F-score (+0.015). \\

Table \ref{compare-fusion-functions} also provides results indicating benefits from unsupervised pre-training.
In RADCURE most models AUC improve after applying pre-training.
This finding does not translate however on the external cohort; which may be explained by the fact that pre-training is performed on the training cohort data.

\section{Conclusion}

In this work, we benchmarked four 3D vision foundation models (VFMs) for 2-year recurrence prediction in head and neck cancers from computed tomography images.
We included three modality fusion functions, to merge clinical and imaging features, and two training strategies.
Our results show that no single VFM dominates across all settings: VISTA3D achieves the best internal performances, while CT-CLIP features demonstrated better generalization on the external cohort.
As an attempt to explain the superiority of CT-CLIP features generalization on the external cohort, one may notice that it is the only imaging foundation model that was pre-trained via an image-text contrastive task.
Its superior performance could suggests that text-derived signals during self-supervised training, help the image encoder learns imaging features that are better at generalizing beyond its pre-training data.
We leave this as an open question for further research.
Additionally, we investigate the choice of the fusion function.
Gating mechanism, inspired by recurrent neural network, showed improved prediction compared to other fusion functions (AUC +0.078 and +0.095 over concatenation and attention respectively on HN-PETCT).
Additionally, we observed that unsupervised adaptation, in the form of few-shot learning pre-training, helps generalization.
We propose a variant of an existing few-shot learning paradigm, that yields the best pre-training strategy results in three out of four VFMs.
Despite its wide adoption, we found the ROC AUC metric insufficient for comparing and clearly identifying the best predictive models when dealing with imbalanced datasets.
Instead, the addition of balanced accuracy (i.e., average of specificity and sensitivity) and F1-score was crucial to assess the capacity of each model to correctly identify patients at risk of recurrence while minimizing the rate of false positives.
Finally, this benchmark therefore identifies the main barriers to leveraging large-scale pre-trained models for individualized risk profiling in head and neck cancer.

\begin{credits}
\subsubsection{\ackname} This study was funded by the AAP SEQ-RTH22 from the French National Cancer Institute.

\subsubsection{\discintname}
The authors have no competing interests to declare that are relevant to the content of this article.
\end{credits}

\bibliographystyle{splncs04}
\bibliography{mybibliography}

\end{document}